\documentclass[letterpaper, 10 pt, conference]{ieeeconf}  % Comment this line out if you need a4paper

\IEEEoverridecommandlockouts                              % This command is only needed if 
\usepackage{graphicx} 
\usepackage{subfigure}
\usepackage{booktabs}
\usepackage{mathtools}
\usepackage{verbatim}
\usepackage{hyperref}
\usepackage{makecell}
\usepackage{amssymb}
\usepackage{multirow}
\usepackage{stfloats}
\usepackage{color}
\usepackage{isotope}
\usepackage{cite}

\title{\LARGE \bf
CNN-based Ego-Motion Estimation for Fast MAV Maneuvers
}

\author{Yingfu Xu and Guido C.\thinspace H.\thinspace E. de Croon% <-this % stops a space
\thanks{The authors are with the Micro Air Vehicle Laboratory, Faculty of Aerospace Engineering, Delft University of Technology, The Netherlands. (emails: \href{mailto:y.xu-6@tudelft.nl}{\texttt{y.xu-6@tudelft.nl}}; \href{mailto:G.C.H.E.deCroon@tudelft.nl}{\texttt{G.C.H.E.deCroon@tudelft.nl}}).
}
}

\begin{document}

\maketitle
\thispagestyle{empty}
\pagestyle{empty}

%%%%%%%%%%%%%%%%%%%%%%%%%%%%%%%%%%%%%%%%%%%%%%%%%%%%%%%%%%%%%%%%%%%%%%%%%%%%%%%%
\begin{abstract}
In the field of visual ego-motion estimation for Micro Air Vehicles (MAVs), fast maneuvers stay challenging mainly because of the big visual disparity and motion blur. In the pursuit of higher robustness, we study convolutional neural networks (CNNs) that predict the relative pose between subsequent images from a fast-moving monocular camera facing a planar scene. Aided by the Inertial Measurement Unit (IMU), we mainly focus on translational motion. The networks we study have similar small model sizes (around 1.35MB) and high inference speeds (around 10 milliseconds on a mobile GPU). Images for training and testing have realistic motion blur. Departing from a network framework that iteratively warps the first image to match the second with cascaded network blocks, we study different network architectures and training strategies. Simulated datasets and a self-collected MAV flight dataset are used for evaluation. The proposed setup shows better accuracy over existing networks and traditional feature-point-based methods during fast maneuvers. Moreover, self-supervised learning outperforms supervised learning. Videos and open-sourced code are available at \url{ https://github.com/tudelft/PoseNet_Planar}

\end{abstract}

%%%%%%%%%%%%%%%%%%%%%%%%%%%%%%%%%%%%%%%%%%%%%%%%%%%%%%%%%%%%%%%%%%%%%%%%%%%%%%%%
\section{INTRODUCTION}

Indoor flight of Micro Air Vehicles (MAVs) is an attractive but challenging task. Towards the goal of autonomy, robust state estimation is one of the most essential modules of the MAV's flight control system. A camera captures rich information in a big field of view. Since being small and power-efficient, it is an ideal onboard sensor~\cite{de2018challenges}. Its combination with the high-sample-frequency IMU is not only suitable for environment perception but also for real-time ego-motion estimation. Visual~\cite{forster2014svo, mur2015orb} and visual-inertial~\cite{li2013high, bloesch2015robust, qin2018vins, sun2018robust, campos2020orb} odometry (VO/VIO) systems contribute to MAVs' autonomy in generic environments by achieving real-time efficiency on onboard processors with decent accuracy. 

\begin{figure}[t]
%	\centering
	\makebox{
		\includegraphics[scale=0.253]{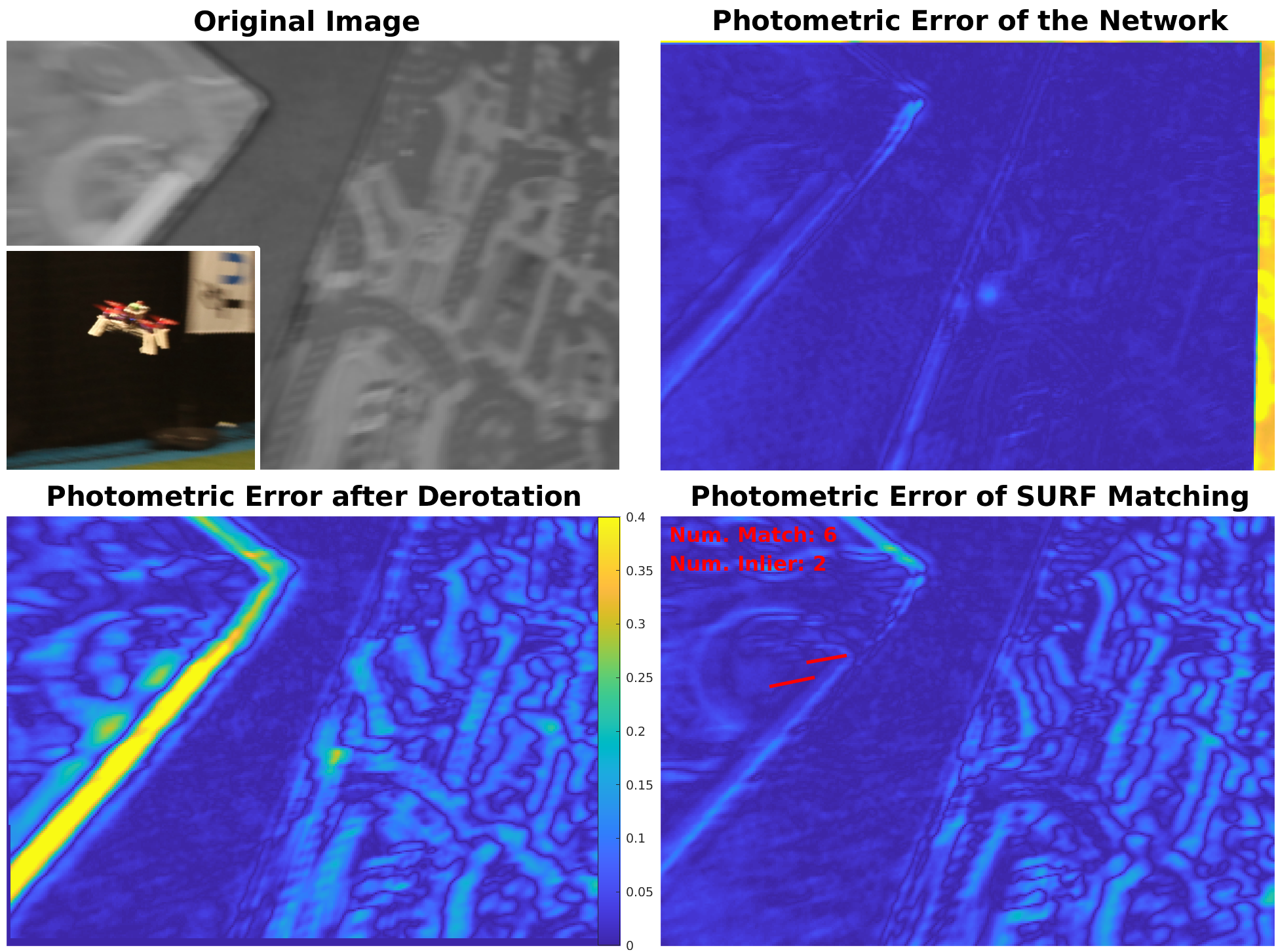}}
	\caption{We study CNNs for visual ego-motion estimation, using blurry gray-scale images captured by the downward-facing camera of fast-moving MAVs. The derotated image pair has big photometric errors. After being warped by the network's prediction of the relative pose, only small photometric errors can be found around edges. The proposed networks better cope with fast motion than traditional feature-based methods.}
	\label{fig1}
\end{figure}

Being constrained by the limited battery life, increasing flight speed is a direct way to enlarge an MAV's operation range and efficiency. However, it also introduces challenges for perception, and notably VO/VIO. 
Detection and tracking of handcrafted interest-point-based features~\cite{shi1994good, trajkovic1998fast, rublee2011orb} is the standard in state-of-the-art VIO systems~\cite{qin2018vins, sun2018robust, campos2020orb}. However, such systems lack robustness in the presence of motion blur occurring during fast maneuvers.
Robust Visual Inertial Odometry (ROVIO)~\cite{bloesch2015robust} directly uses photometric errors of multilevel image patches around FAST feature points~\cite{trajkovic1998fast} to be more robust against image blur than point features, partly because the texture of the tracked image patch is taken into account. However, Foehn \textsl{et al.} point out that, at larger speeds the state estimation of ROVIO suffers from drift~\cite{foehn2020alphapilot}. When the speed gets larger, feature points and patches can move out of the camera's field of view sooner. We believe that the bigger visual disparities between images and the consequent lower number of frames in which features can be tracked is another adverse condition besides motion blur. Since ROVIO takes features' 3-dimensional (3-d) positions as states of an extended Kalman filter (EKF), having fewer visual observations decreases the accuracy. Other VIO systems such as~\cite{li2013high, qin2018vins} that estimate feature positions by multiple observations can also suffer from high-speed motion~\cite{zhong2020direct}.

CNNs are state of the art in many computer vision tasks and are promising for VO as well. Various networks have been proposed to estimate the pose change (rotation and translation) between two or more subsequent views. There are not only supervised pose networks trained by limited ground truth~\cite{costante2015exploring,melekhov2017relative,wang2018end} but also self-supervised ones trained together with other networks including a depth estimation network~\cite{zhou2017unsupervised,li2018undeepvo,ranjan2019competitive,chen2019self,bian2019unsupervised}. Evaluated by the KITTI dataset~\cite{geiger2012we}, these networks obtain highly accurate performances and rival VO~\cite{mur2015orb} with a traditional vision method~\cite{rublee2011orb}.

There are also pose networks considering the application to an MAV's ego-motion estimation. For example, in~\cite{wang2018end}, a recurrent CNN is trained on the EuRoC MAV dataset~\cite{burri2016euroc} to regress the 6 degree-of-freedom (6DoF) motion. The network manages to learn the more complex (compared with a car) MAV's dynamics but the accuracy is limited by the small amount of training data. Differently, PRGFlow~\cite{sanket2020prgflow} focuses on the essential function of estimating the 3-d translational velocity of the MAV with a downward-facing camera, assuming a planar ground. Aided by the attitude estimated by IMU measurements, via image warping, the task is simplified to the pixel-level similarity transformation estimation. Although PRGFlow thoroughly studied CNN-based ego-motion estimation, the focus is on the low-speed flight (about 0.5m/s on average), with motion blur lacking from the artificially generated training images. 

Hence, it is currently still an open question of how good CNNs perform during fast maneuvers. 
%Towards the consequential adverse conditions, an intuitive hypothesis is that image patches can look similar in blurry images that are time adjacent, and thus convolutional kernels in multiple layers can learn to capture the correspondences even if big visual disparities exist. 
To gain insight into this matter, in this article we study networks predicting 3-d relative translation of MAVs in fast maneuvers with a downward-facing camera. The networks are trained and tested on images with significant motion blur and big visual disparities. Our main contributions are that we: (1) Extend and further improve the performance of the network framework proposed in~\cite{sanket2020prgflow} to fit fast maneuvers, and (2) Investigate how well the networks can deal with faster motion in comparison with traditional feature-point-based methods.
%, and {\color{blue}(3) Integrate the network into an onboard visual-inertial ego-motion estimator that navigates an autonomous MAV flying at high speed.}
According to our knowledge, this is the first work showing networks' superior performance in fast motion when traditional feature-point-based methods have high failure rates.

\section{METHODOLOGY}\label{section:METHODOLOGY}

\subsection{Homography Transformation}

As shown in Eq. \ref{eq1}, for a fixed point laying on a plane observed by two cameras, it has been proven in~\cite{faugeras1988motion} that the projective coordinates $\boldsymbol{x_{1}}$, $\boldsymbol{x}_{2}$ of the same point in the camera frames are related by the homography matrix $\boldsymbol{H}$ that depends only on the 6-d relative pose of the cameras and the unit normal vector of the plane $\boldsymbol{n}$. $\boldsymbol{R}$ denotes the rotation matrix between the camera frames and $\boldsymbol{t}$ denotes the translation vector expressed in the second camera's frame pointing from the second camera to the first one. The scalar $d$ is the distance from the first camera to the plane. 
% t in cam2 frame and n in cam1 frame

\begin{equation}\label{eq1}
\boldsymbol{x}_{2}= \boldsymbol{H}  \boldsymbol{x}_{1},\ 
\boldsymbol{H}=\boldsymbol{R}+\frac{\boldsymbol{t} \boldsymbol{n}^{T}}{d}
\end{equation}

Here we define the coordinate system whose $x$-axis points to the north, $y$-axis to the east, and $z$-axis to the gravity direction as the world frame. The plane that the downward-facing camera observes is assumed to be orthogonal to the gravity vector. The attitude of the camera relative to the world frame can be estimated by an IMU, then the information remaining unknown in the homography matrix is the ratio of the translation vector $\boldsymbol{t}$ to the distance to the plane $d$. Here we refer to it as the distance-scaled relative translation vector. This vector together with the flight height that is available from a downward-facing rangefinder can determine the metric average translational velocity of the MAV during the camera's sample interval. Here we refer to it as the distance-scaled translational velocity vector.

PRGFlow warps both the images to make the image planes parallel to the ground using the absolute attitude estimated by the IMU. The distance-scaled relative translation then can be determined by the similarity transformation between the image pair. Networks are trained to predict the 3 parameters (2-d translation, zoom-in/out) reflecting the relative location of pixels. However, when the roll or pitch angle of the MAV is big, which is often the case in fast maneuvers, the camera would have big tilt angles relative to the plane's normal vector. So warping the image pair like PRGFlow can cause big black boundaries and thus lose many pixels. It then requires pre-processing moving the pixels back inside the image frame and the corresponding post-processing for calculating the pose from the similarity transformation. 

To avoid the above-mentioned processings, our networks predict the distance-scaled relative translation vector expressed in the camera frame directly from images that have (non-zero) tilt angles, requiring input images to have the identical intrinsic parameters as the training set. Only one image needs to be warped by the relative rotation. Tilt angles are available from an IMU but we additionally explore networks predicting them in subsection \ref{section:tilt}.

\begin{figure}[b]
	\centering
	\makebox{
		\includegraphics[scale=0.39]{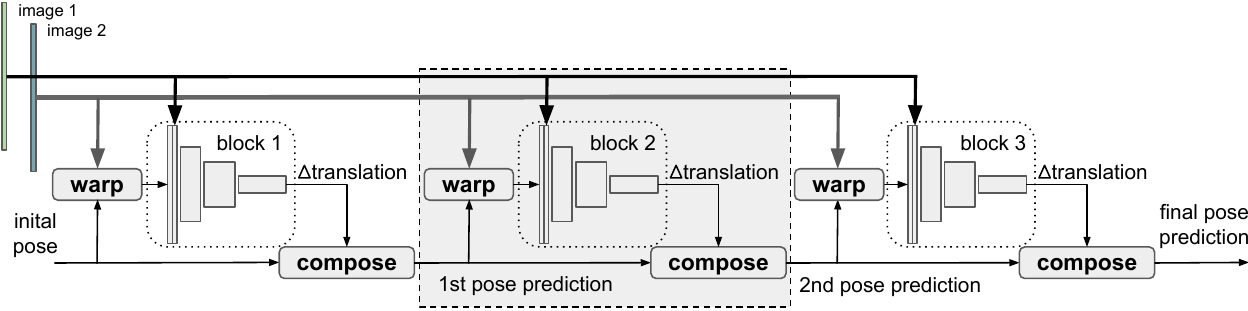}}
	\caption{An ICSTN-based network with 3 blocks for relative pose prediction. The dashed line frame indicates the basic functional unit that can be sequentially stacked one or multiple times. The dotted frame indicates a network block that takes (downsampled) concatenated images as input and predicts the 3-d distance-scaled relative translation.}
	\label{icstnFig}
\end{figure}

\subsection{Cascaded Network Blocks Connected by Image Warping}

Sanket \textsl{et al.} adopt the inverse compositional spatial transformer networks (ICSTN)~\cite{lin2017inverse} as the framework of their networks~\cite{sanket2020prgflow}. The ICSTN has multiple network blocks that predict the image deformation that benefits the final goal. 
%The image is warped by the output of one network block before input to the next block. 
Based on the homography transformation, a new image can be synthesized by warping the original image using the method proposed in~\cite{jaderberg2015spatial}. As shown in Fig. \ref{icstnFig}, a network block is made up of multiple convolutional layers followed by a fully-connected layer regressing the translation. With multiple cascaded network blocks, each block takes the concatenated original image 1 and the image 2 warped by the newest pose prediction as input and combine its output into the pose prediction. As the pose prediction is refined by more blocks, there is less relative motion between the concatenated images. Each block predicts a part of the total relative translation, making the problem more tractable. 
%\textbf{TODO} \cite{erlik2017homography} use similar network architecture for homography transformation estimation. 
The network can also make use of an initial guess of the relative pose, which can be available from the IMU integration or the MAV's dynamic model. PRGFlow has compared network architectures inside one block. Focusing on fast maneuvers, we study higher-level architectures applying to the pyramidal images and feature maps to enlarge the receptive field which is important for dealing with the big visual disparities. 

The loss functions of the networks are the mean of the Charbonnier~\cite{sun2014quantitative} loss of the predicted 3-d translation's error in supervised learning and the mean of the Charbonnier loss of the valid pixels' photometric error in self-supervised learning. For data augmentation, we feed the network with image pairs concatenated in both orders to perform bidirectional training.

We implement the networks in Python 3.6.9 with the Pytorch~\cite{paszke2019pytorch} 1.1.0 library. The Adam optimizer~\cite{kingma2014adam} with $\beta=(0.9, 0.999)$ is utilized during the 25 training epochs. The batch size is 16. The initial learning rate is $0.0002$ and it is divided by 2 after 5, 10, 15, and 20 epochs. The weights of convolutional layers are initialized by Glorot initialization~\cite{glorot2010understanding} with a gain of 1. The weights of fully-connected layers are initialized by the (Pytorch default) uniform distribution $\mathcal{U}(-\sqrt{k}, \sqrt{k})$ where $k$ is the multiplicative inverse of the number of input features. 

\subsection{Dataset Generation} \label{section:DatasetGeneration}
We use the Microsoft COCO dataset~\cite{lin2014microsoft} as the source of a large variety of textures to generate a big number of image pairs thanks to the homography transformation. A source image is treated as a plane above which a simulated camera is moving. For one plane, one image pair is generated. Costante \textsl{et al.} and Kendall \textsl{et al.} test their pose estimation networks with respectively artificial Gaussian blur~\cite{costante2015exploring} and motion blur~\cite{kendall2015posenet} added to images. Their blur is uniform over the whole image and thus not ego-motion-related. In order to obtain realistic blur that is caused by the camera's motion within the exposure duration, we simulate a moving camera whose time step for the kinetic integration is 0.1 millisecond (ms) and the exposure duration is 10ms. In each integration step during the exposure, an image is sampled from the homography transformation of the plane. The blurry image is the average of the 100 sampled images. The poses of both the exposure starting step and ending step of an image are recorded. Except for subsection \ref{section:selfSupervised}, the pose of the starting step is used as the ground truth in supervised learning. 30 frames per second (fps) are recoreded to simulate a common global shutter camera. The images are in grayscale with the resolution of $320\times224$ pixels. The intrinsics of the simulated camera are $f_x=160$, $f_y=160$, $c_x=160$, $c_y=112$. 

\begin{figure}[t]
	\centering
	\makebox{
		\includegraphics[scale=0.29]{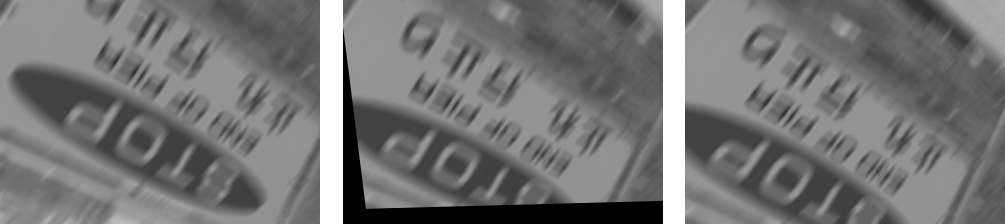}}
	\caption{A blurry image pair. Starting from the left: (1) the first image, (2) the derotated second image, and (3) the second image.}
	\label{fig2}
\end{figure}

The initial poses of the kinetic integrations uniformly distribute within a normal quadrotor MAV's flight envelope. The uniformly randomly generated translational velocity and rotational velocity stay constant during the kinetic integration. The camera's distance-scaled translational velocity vector's components along the $x$-axis and $y$-axis of the world frame range from -7.5 to 7.5. The range of the component along the $z$-axis is from -3.75 to 3.75. Angular velocity vector's components along the $x$-axis and $y$-axis of the camera frame range from -180 to 180 degrees per second. The range of the component along the $z$-axis is from -90 to 90 degrees per second. Initial roll and pitch angles range from -25 to 25 degrees. Since we record the poses at the start and the end of exposure duration, the motion flow that causes blur can be calculated. Over the dataset, the average motion flow of all the pixels in an image has mean values of 6.6 and 6.2 pixels in the $x$-axis and $y$-axis, respectively. The maximum motion flow of all the pixels in an image has mean values of 13.4 ($x$-axis) and 11.9 ($y$-axis) pixels. The above data shows that our dataset involves a big range of motion and significant motion blur. After removing hundreds of images with little texture, there are 82,172 training samples, 9,948 validation samples (for validating the model after each epoch during training), and 30,565 testing samples.

\begin{table*}[t]

	\caption{ICSTN-based Networks with Different Numbers of Blocks}
	\centering
	\label{icstn}
	% width 391
	\begin{tabular}  {p{26pt}<{\centering}p{13pt}<{\centering}p{46pt}<{\centering}p{17pt}<{\centering}p{10pt}<{\centering}p{8pt}<{\centering}p{17pt}<{\centering}p{118pt}<{\centering}p{136pt}<{\centering}} 
		\toprule  
		Network& Num. Blocks& Num. Conv./ Kernel/ Stride& Num. Params& FPS& RF& Inlier Rate($\%$)& EPE's Standard Deviations (1e-3) (end-point loss / multi-stage losses) &Medians of EPE's Absolute Values (1e-3) (end-point loss / multi-stage losses)\\
		\midrule  
		1 \cite{zhou2017unsupervised}& 1& 8/ 7,5/ 2.2& 1.583M& 215& 263& 90.14&(13.44,13.88,19.23)&(7.83,8.11,11.93)\\ % Godzilla1  \textsc{10-15-17:23} 0.0266 SfMLearner~\cite{zhou2017unsupervised}
		%		VanillaNet& 12& 1.342M& 163& 895&0.0166\\
		
		2 \cite{huang2017densely}& 1& 18/ 3,3/ 2,2& 1.441M& 105& 759&91.77&(6.70,6.94,9.63)&(3.85,4.03,6.12) \\ % godzilla0  10-15-23:33  DenseNet~\cite{huang2017densely}
		
		3 \cite{he2016identity}& 1& 21/ 3,3/ 2,2& 1.477M& 103& 975&92.35&(7.32,7.58,10.57)&(4.33,4.48,6.66)\\ % mavlabgpu1  10-15-20:17 0.0143 ResNet~\cite{he2016identity}
		
		4 (ours)& 2& 9/ 7,5/ 2,2& 1.385M& 104& 647& 89.92&(3.35,3.30,3.91) / (2.71,2.64,3.28)& (1.89,1.89,2.61) / (1.49,1.49,2.18)\\ %godzilla 10-16-00:27 0.0051
		%		2& 9/ 3,3/ 2,2& 1.385M& 104& 647& 0.0; 1.0& 89.53& (0.07,-1.47,7.13)/ (3.35,3.25,3.87)&(3.17,3.14,3.59)/ (1.91,1.87,2.59)\\ %mavlabgpu0  10-16-17:15  0.0063
		5 (ours)&3& 5/ 7,5/ 2,2& 1.367M& 101&  71&  87.59& (3.33,3.15,4.29) / (2.28,2.22,3.08)& (1.91,1.83,2.90) / (1.24,1.22,2.11)\\ %mavlab 10-15-23:38 0.0058/ 
		%		3& 5/ 7,5/ 2,2& 1.367M& 101&  71& 0.0; 0.0; 1.0& 82.32& (-0.69,-1.30,7.92)/ (3.35,3.19,4.32)&(4.19,3.97,4.90)/ (1.92,1.83,2.90)\\ %mavlabgpu1  10-16-17:25  0.0087
		
		6 (ours)&3& 5/ 7,5/ 4,2& 1.252M& 101& 135& 92.18& \textbf{(2.06,2.05,2.90)}& \textbf{(1.16,1.13,2.00)}\\ % godzilla0  10-16-17:33 re-run
		%		3& 5.7454& M& [0.2,0.3,0.5]& 247& 101& 0.?\\
		
		7 (ours)&4& 3/ 7,5/ 4,4& 1.421M& 95& 55&  85.48& (3.58,3.51,4.75) / (2.20,2.19,3.06)& (2.20,2.20,3.20) / (1.28,1.26,2.13)\\ % godzilla1 10-17-23:20 0.0067/ 
		%		4& 3/ 7,5/ 4,4& 1.421M& 95& 55& 0.0; 0.0; 0.0; 1.0& 76.83& (1.26,1.98,6.29)/ (3.61,3.53,4.79)&(5.84,5.79,6.22)/ (2.21,2.20,3.24)\\ % godzilla0 10-17-23:19 0.0119
		
		8 (ours)&4& 3/ 9,5/ 8,4& 1.276M& 95& 105& 87.35& (2.10,2.09,3.01)&(1.25,1.22,2.09)\\ % godzilla2 10-17-23:35

		\bottomrule %
	\end{tabular}
	
\end{table*}

\begin{table*}[t] % !htbp
	
	\caption{ICSTN-based Networks using Pyramidal Images or Feature Maps} %  with different number of blocks
	\label{icstnPyramid}
	\centering
	\begin{tabular} {p{28pt}<{\centering}p{148pt}<{\centering}p{16pt}<{\centering}p{46pt}<{\centering}p{78pt}<{\centering}p{111pt}<{\centering}}%{c{20pt}c{20pt}c{20pt}c{20pt}c{20pt}c{20pt}}
		\toprule  
		Num. Pyramid& Num. Layers/ Kernel/ Stride& Num. Params& FPS (intrpl. / avg. pooling)& Receptive Field& EPE's Standard Deviations (1e-3) (intrpl. / avg. pooling)\\
		\midrule  
		%			2& 9& 1.385M& [0.3,0.7]& 99& 0.0210\\

		3& 3/ 7,5/ 2,2; 4/ 7,5/ 2,2; 5/ 7,5/ 2,2& 1.388M& 103 / 108& 23$\times$4; 39$\times$2; 71& (2.13,2.05,2.93) / (2.11,2.09,2.91)\\ % mavlab0 10-18-01:36 0.0051 /  mavlab0 10-19-11:48 0.0051 
		3& 3/ 7,5/ 4,2; 4/ 7,5/ 4,2; 5/ 7,5/ 4,2& 1.272M& 104 / 111& 39$\times$4; 71$\times$2; 135& (2.11,2.09,2.94) / (2.09,2.06,2.94)\\ % mavlab1 10-18-02:00 0.0049 / mavlab1 10-18-14:40 0.0048
		%			3& [4,4.54,4.7454]& 1.245M& [31*4,61*2,119]& 102& 0.?\\
		3& 4/ 7,5/ 2,2; 4/ 7,5/ 4,2; 4/ 7,5/ 4,4& 1.316M& 104 / 109& 39$\times$4; 71$\times$2; 119& (2.07,2.06,2.93) / (2.07,2.07,2.90)\\ % mavlab0 10-18-14:52 0.0047? godzilla0 10-21-17:52 0.0047/ godzilla0 10-18-14:55 0.0047
		4& 2/ 7,5/ 2,2; 2/ 7,5/ 4,2; 3/ 7,5/ 4,2; 3/ 7,5/ 4,4& 1.386M& 93 / 98& 15$\times$8; 23$\times$4; 39$\times$2; 55& (2.03,2.02,2.87) / (2.02,2.02,\textbf{2.85})\\ % godzilla1 10-18-15:27 0.0054 / godzilla2 10-18-15:30 0.0055
		3& [FPE: 3/ 7,5/ 2,2] + [2; 3; 4]& 1.455M& 100& 71; 71; 71& (2.18,2.12,3.12)\\ % PWC godzilla2  10-15-19:19  0.0054
		3& [FPE: 3/ 7,5/ 4,2] + [2; 3; 4]& 1.340M& 100& 135; 135; 135& (\textbf{2.00},\textbf{1.97},3.08)\\ % mavlab0 10-18-15:51  0.0047
		\bottomrule 
	\end{tabular}
	
\end{table*}

\begin{table*}[t] % !htbp
	\centering
	
	\caption{Comparison between Supervised and Self-Supervised Learning}
	\label{self-sup}
	\centering
	\begin{tabular} {p{52pt}<{\centering}p{100pt}<{\centering}p{147pt}<{\centering}p{152pt}<{\centering}}
		\toprule  
		Networks& Inlier Rate($\%$) (supervised / self-supervised)& EPE's Standard Deviations (\textbf{1e-4}) (supervised / self-supervised)& Medians of EPE's Absolute Values (\textbf{1e-4}) (supervised / self-supervised)\\
		\midrule

		Table \ref{icstn} 6&  92.18 / 91.41& (20.62, 20.46, 28.99) / (19.16, 19.29, 28.36)& (11.62, 11.29, 20.01) / (10.71, 10.60, 19.54)\\ % same as table I / mavlab0 12-04-20:43 self-supervised
		
		Table \ref{icstn} 6*&  75.20 / 70.84& (5.30, 4.95, 7.04) / (4.16, 3.82, 5.54)& (3.15, 3.15, 4.55) / (2.40, 2.15, 3.47)\\ % 12-10-01:19 mavlab1 / self-supervised 12-04-20:40 godzilla1
		% train set (std,1e-3) (1.85, 1.88, 2.81) / (1.85, 1.87, 2.77)
		
		%		mavlab1 0.0024 10-21-18:01 
		%		Table \ref{icstn} 6*& Self-Supervised& 67.32& (-0.08, 3.89, -3.77)/ (5.05, 4.39, 6.38)& (2.21, 2.00, 1.83)/ (2.78, 2.54, 4.10)\\ % self mavlab0 0.0040 10-21-17:59
		%		Table \ref{VanillaNets} 3*& Supervised& 91.52& (8.31, 49.20, -35.04)/ (62.95, 65.70, 92.39)& (5.27, 5.51, 8.13)/ (36.26, 37.69, 59.09)\\ % 10-22-19:01 mavlab0 
		%		Table \ref{VanillaNets} 3*& Self-Supervised& 88.29& (4.35, -3.88, -39.67)/ (67.65, 66.94, 96.86)& (6.07, 5.92, 8.78)/ (37.71, 37.11, 60.47)\\ % 10-22-19:03 mavlab1
		Table \ref{icstnPyramid} 3(p)* &  73.20 / 69.91&(5.49, 4.93, 7.02) / (4.20, 3.79, 5.52)&(3.08, 2.91, 4.42) / (2.36, 2.16, 3.47)\\ %  Supervised 10-22-19:18  Godzilla1 avg.pooling
		%		Table \ref{icstnPyramid} 3(p)*& Self-Supervised& 69.68& (1.48, -3.08, 2.28)/ \textbf{(4.26, 3.92, 5.46)}& (1.39, 1.31, 1.18)/ \textbf{(2.44, 2.23, 3.46)}\\ % 10-22-19:16 Godzilla0 avg.pooling
		% train set (std,1e-3) (1.84, 1.85, 2.78) / (1.85, 1.85, 2.75)

		Table \ref{icstnPyramid} 4(i)*& 71.43 / 67.29& (4.49, 4.23, 5.94) / \textbf{(3.08, 2.89, 3.90)}& (2.82, 2.64, 3.93) / \textbf{(1.91, 1.80, 2.63)}\\ % Supervised 10-23-15:25 mavlab1
		%		Table \ref{icstnPyramid} 4(i)*& Self-Supervised& 66.90& (-7.60, -6.71, -5.43)/ \textbf{(3.03, 2.88, 3.86)}& (2.14, 1.96, 1.68)/ \textbf{(1.91, 1.79, 2.61)}\\ % 10-23-15:24 godzilla0 
		%		Table \ref{icstnPyramid} 5*& Supervised& 74.08& (-2.49, 4.69, -4.58)/ (7.83, 7.42, 10.75)& (1.64, 1.54, 1.84)/ (4.62, 4.50, 6.92)\\ % 10-26-12:39 godzilla1
		%		Table \ref{icstnPyramid} 5*& Self-Supervised& 69.82& (1.11, -1.07, -3.07)/ (5.97, 5.72, 8.28)& (2.03, 1.87, 1.93)/ (3.36, 3.10, 5.37)\\ % 10-26-12:32 mavlab1
		
		\bottomrule
	\end{tabular}
	
\end{table*}

\section{NETWORKS}\label{section:NETWORKS}

\subsection{ICSTN-based Networks}\label{section:Cascaded Network Blocks} 
In this subsection, we study ICSTN-based networks with different numbers of blocks. Blocks of one multi-block network have identical architecture. In the 3rd column of Table \ref{icstn}, ``Num. Conv." is the abbreviation of the number of convolutional layers. The kernel sizes and the strides of the first and second convolutional layers are also listed. Deeper layers have kernel sizes of 3 and strides of 2 or 1. Networks' inference speeds are indicated by fps when running on an NVIDIA Jetson TX2 with Ubuntu 18.04.3 LTS and Cuda V10.0.326 in the MAXP\_CORE\_ARM power mode.
%that has the highest inference speed of our implementation. 
``RF" denotes the receptive field of the fully-connected layer's input. We call the final prediction error the end-point error (EPE). The predicted translation is rotated into the world frame to calculate the 3-d EPE vector.
%Since a small number of images with little texture exists in the dataset, 

EPE's standard deviation can reflect how noisy the predictions are but it is sensitive to outliers that can be caused by image pairs lacking texture or having duplicate textures. So we use a local outlier rejection function of MATLAB to remove outliers and keep the characteristic of local distribution. After respectively ascendingly sorted by the corresponded ground truth along each axis, the EPEs whose absolute values are more than 3 scaled median absolute deviations in a local window of size 1000 are rejected. A prediction is considered an outlier if its component along any axis is rejected.
The 3 values inside the bracket separated by commas correspond to the data of the $x$-axis, $y$-axis, and $z$-axis. The medians of EPE's absolute values are calculated from all the predictions including outliers.

%Except that the network block with 9 convolutional layers has 2 layers that have the stride of 1, all the other convolutional layers after the second layer of the networks in this table have the kernel size of 3 and the stride of 2. 

Networks with a single block are shown in the first 3 rows of Table \ref{icstn}. The 1st network is the pose estimation network proposed in~\cite{zhou2017unsupervised}. The 2nd and 3rd networks respectively have skip connections~\cite{huang2017densely,he2016identity} for better performance in their deeper architectures.
%All the convolutional kernel has a size of 3 in the 2 deeper networks. 7 layers have the stride of 2 and other layers of 1 in both deep networks.
They have smaller model sizes, higher accuracy, but slower speed. The 2nd network with 18 convolutional layers and densely connected architecture has the highest accuracy. In the case of multiple blocks, for each block, there is an image warping operation that has unneglectable time consuming. Since a network's inference speed is required to be around 100fps, the total number of layers in the whole network decreases when the number of blocks increases.
%When there are 4 blocks with only 3 convolutional layers and a fully-connected layer in each block, the inference speed cannot reach 100Hz. 
The accuracy gets worse when there are more than 3 blocks mainly because the blocks are too shallow and the total capacity of the whole network decreases. As shown by the 6th and 8th networks, bigger strides lead to smaller resolution of the last feature map and thus fewer parameters in the fully-connected layer. Besides, it increases the receptive field. Since our dataset has image pairs with big visual disparities, a bigger receptive field can capture more feature correspondences and improve the accuracy.

For the networks with multiple blocks, instead of only using the loss of the final prediction (end-point loss) in training~\cite{sanket2020prgflow}, we weighted sum the losses of every prediction after every block (multi-stage losses) for backpropagation. The loss weight distributions of blocks are respectively [0.3,0.7], [0.2,0.3,0.5], and [0.1,0.2,0.3,0.4] for networks with 2, 3, and 4 blocks. The accuracy is compared by the 4th, 5th, and 7th networks of Table \ref{icstn}. Multi-stage losses produce higher accuracy. For the 6th and 8th networks, we only show the results of multi-stage losses. All the networks in the rest part of this article are trained with their multi-stage losses.

\subsection{Pyramidal Images and Feature Maps in ICSTN}\label{section:pyramid}

From Table \ref{icstn}, we find that a bigger receptive field can benefit accuracy. When the kernel size and stride keep the same, another way to increase the receptive field is using pyramidal images. Networks using pyramidal images or feature maps with lower resolution are shown in Table \ref{icstnPyramid}. The downsampled image at each pyramid level has half the size of the image of its adjacent lower level. So the lowest-resolution image of the network that has 4 pyramids has one-eighth the width and height of the original image. The number of network blocks is the same as the number of pyramids. The first pose prediction block uses the images at the highest pyramid level with the lowest resolution. The predicted pose is used to warp the original image. Then the warped image is downsampled to the next lower pyramid level and input to the next network block. For image downsampling, we compared bilinear image interpolation~\cite{jaderberg2015spatial} and average pooling. They have similar accuracy, but average pooling is faster in our Pytorch implementation. 

Since the EPE's standard deviation is enough to reflect the accuracy of the network predictions, the medians of EPE's absolute values are not shown in Table \ref{icstnPyramid}. Comparing the 1st network of Table \ref{icstnPyramid} with the 5th of Table \ref{icstn}, with the same kernel size and stride, the pyramidal network that has fewer layers achieves higher inference speed and accuracy, thanks to the bigger receptive fields of the first two blocks. 
%The 4th network in Table \ref{icstnPyramid} and the 7th network in Table \ref{icstn} show the same result for 4-block networks. 
Comparing the 2nd network of Table \ref{icstnPyramid} with the 6th network of Table \ref{icstn}, the pyramidal network has slightly lower accuracy. We think it is because when the receptive fields are big enough, the pyramidal version receives less information due to the downsampling. The 8th network of Table \ref{icstn} has a big receptive field. Also with 4 blocks, the 4th network of Table \ref{icstnPyramid} has decreasing receptive fields with the increasing of image resolution. Although 3 out of 4 blocks have smaller receptive fields than the 8th network of Table \ref{icstn}, this 4-stage coarse-to-fine refinement gets better accuracy. The 2nd and 3rd networks of Table \ref{icstnPyramid} have the same total number of layers. The 3rd one having a deeper block at the lowest resolution achieves slightly higher accuracy.

%\subsection{Feature Pyramid Extractor Networks}\label{section:Extractor}

The pyramidal feature maps network is based on the feature pyramid extractor (FPE) inspired by the PWC-Net~\cite{sun2018pwc}.  The results are shown in the last 2 rows of Table \ref{icstnPyramid}. The general principle is extracting multiple feature maps at different resolutions (pyramid levels) of each image respectively by the same convolutional feature extractor network. One of the feature maps is warped and then concatenated along the channel dimension with the other feature map of the same size. The concatenated feature maps are the input of the pose prediction blocks. The networks we design have 3 levels of pyramidal feature maps and 3 pose prediction blocks that have 2, 3, and 4 convolutional layers respectively. The FPE network at the last row of Table \ref{self-sup} has the highest accuracy in the $x$-axis and $y$-axis.
% but the error in the $z$-axis is relatively bigger.

\subsection{Self-Supervised Learning}\label{section:selfSupervised}

Self-supervised learning is based on the photometric error between the image warped by the predicted relative pose and the other image. We use a mask to not count the photometric errors of the pixels whose locations to interpolate lie outside the image frame. 
%(note that the size factor of data in the column title of this table is 1e-4 instead of 1e-3 in previous tables)
By the results shown in the 1st row of Table \ref{self-sup}, we notice that self-supervised learning with a basic photometric loss gets better accuracy than supervised learning. The reason behind it worth further studying. For now, we think it is mainly because the target relative poses used in supervised learning are calculated from the poses at the starting time points of the image exposure. While the simulated camera keeps moving within the exposure duration, motion blur appears and the image gets a different appearance from the start of exposure, and thus there will be small photometric errors between the images warped by the target relative pose. 
%The relative pose that has the minimum photometric error should be determined by the poses at some uncertain time point around the middle time point of the exposure duration. 
%So when we train the network with the target pose that does not have the minimum photometric error, 
This means the network is trained to regress to a target not perfectly matching the feature correspondences. This discrepancy can ``confuse" the network. While in the case of self-supervised learning, the network tries to minimize the photometric error affected by the blur and is more likely to converge to the ``accurate" relative pose that best matches the feature correspondence. 
When we evaluate the self-supervised network, we use the poses at the start of exposure as the ground truth, to which the network does not learn to converge. But the effect of it is smaller than the ``confusion" induced by the discrepancy.

To verify the hypothesis above, we use the average of the poses of the start and the end of exposure as the pose of a blurry image and calculate the target relative pose from it. The results are marked with an asterisk and shown in Table \ref{self-sup} from the 2nd to the 4th row. ``Table \ref{icstnPyramid} 3(p)" denotes the average pooling version of the 3rd network of Table \ref{icstnPyramid}. Similarly, the ``(i)" denotes the bilinear interpolation version.
%We highlight the best networks running faster (3rd row) and slower (4th row) than 100Hz.

From the results of the testing set shown in Table \ref{self-sup}, one can notice that all the self-supervised networks are more accurate. Besides, they are also slightly more accurate on the training set. As for the supervised networks,  training with the new target pose (2nd row) has much higher accuracy compared with the old target pose calculated from the poses at the start of exposure (1st row). The inlier rates drop when we use the new target pose. The reason is that the errors of the image pairs having less texture are more likely to be outliers because their neighbors have smaller errors. 
Obviously, the new target pose matches the feature correspondence better and acts as better supervision. But still, the remaining small discrepancy makes it less good than self-supervised networks. 
So we believe that self-supervised learning is a better choice for blurry image pairs that have unknown relative pose perfectly matching the feature correspondences. 
This also provides us with the insight that taking the non-neglectable exposure duration of an image into account 
%to refine its timestamp 
can benefit ego-motion estimation.

\subsection{Networks for Tilt Angle Prediction}\label{section:tilt}

%If the plane is flat in the world frame, the tilt in the view is actually equal to the drone's pitch angle.

It is known that one can estimate the tilt of the camera relative to the plane in the view from the optical flow field~\cite{de2013optic}. 
Since tilt is a property of the flow field and hence affects both images, it cannot be estimated iteratively by our ICSTN-based framework that warps only one image (Fig. \ref{icstnFig}).
For this preliminary investigation, we employ a single deep network block to predict tilt angles from a pair of derotated images, supervised by ground truth. 
%As shown in Fig. \ref{icstnFig}, we only warp one image. Unlike the relative translation, the tilt angles of the image not being warped keep constant and it cannot get smaller by combining the predictions from blocks. So there is no point to have more than one block in the network. We train the networks with one deep block to predict them from a pair of derotated images, supervised by ground truth. 
Shown in Table \ref{tiltnets}, the best network's EPE's standard deviation is around 4 degrees for both angles. Although the prediction is noisy, it may serve as an unbiased absolute information source of attitude. 

\begin{table}[!htbp] % finished test re-run 
	\centering
	\setlength{\tabcolsep}{1.35mm}{
		
		\caption{Networks Predicting Tilt Angles}
		\label{tiltnets}
		\centering
		\begin{tabular} {p{34pt}<{\centering}p{23pt}<{\centering}p{78pt}<{\centering}p{70pt}<{\centering}}%{c{20pt}c{20pt}c{20pt}c{20pt}c{20pt}c{20pt}}
			\toprule  
			Networks& Inlier Rate($\%$)& EPE's Std. Dev. (radian, 1e-2)& Medians of EPE's ABS (radian, 1e-2)\\
			\midrule  
			Table \ref{icstn} 2*& 97.13& \textbf{(7.72, 6.57)}& \textbf{(5.06, 4.30)}\\ % 10-23-13:30 mavlab0 tilt densenet18 
			Table \ref{icstn} 3*& 96.91& (8.71, 7.50)& (5.75, 4.87)\\ % 10-22-23:16 godzilla2 tilt resnet21
			
			\bottomrule 
		\end{tabular}
	}
\end{table}

\section{EVALUATION}\label{section:EVALUATION}

%In this section, we further evaluate the networks facing the real world situation: inaccurate attitude. And compare networks with traditional feature-based methods. Since feature-based methods can using a sequence of images to build a optimized point cloud map that may be much more accurate, in our comparison we only take inputting one image pair into account.
%We use the 4th network of Table \ref{self-sup} for evaluation and comparison to traditional feature-based methods. We use MATLAB functions for feature detection and matching. Note that one can get more points by tuning the parameters of feature detection. Here we only show the results of the default parameters. If enough points are detected, 50 uniformly distributed ones are selected. The translation is obtained by calculating the similarity matrix (with known in-plane rotation, 3 degrees of freedom left) based on linear least squares and random sample consensus (RANSAC).
%NOTE all the data of Table \ref{self-sup} 10 is using original image without initial guess.

The 4th network of Table \ref{self-sup} is chosen for evaluation and comparison to traditional feature-based methods. Note that the network is trained only with the simulation dataset described in subsection \ref{section:DatasetGeneration} without any fine-tuning to highlight the generalizability.
We use MATLAB functions for traditional feature detection and matching. More feature points can be detected by tuning the parameters of the functions. Here we only show the results of the default parameters. 50 uniformly distributed ones are selected when there are enough detected features. The translation is obtained by calculating the similarity matrix (with known in-plane rotation, 3 DoF left) based on linear least squares and random sample consensus (RANSAC).

\subsection{Simulated Dataset}

We generate a dataset of 5000 image pairs with different exposure duration (ranges from 0.2ms to 20ms) and random distance-scaled velocity vectors having the same norm ($\|\boldsymbol{v}\|/d=5$) to compare the performance of the network and feature-based methods with increasing motion blur. Another dataset of 5000 sharp image pairs with different distance-scaled velocity vectors (same range as the training set) and the same exposure duration (0.2ms) is generated to study the effect of visual disparity. All the image pairs have the random attitude and zero angular rates. 

\begin{figure}[!htbp]
	\centering
	\makebox{
		\includegraphics[scale=0.65]{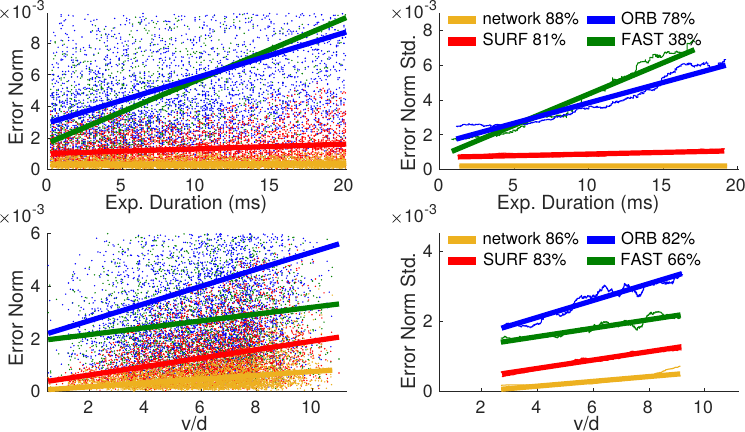}}
	\caption{Comparison between feature-based methods and the network. The top row shows how their accuracy changes with the amount of motion blur. The bottom row shows the effects of the increasing visual disparity. ``v/d" denotes the norm of the distance-scaled velocity of the simulated camera.}
	\label{blurError}
\end{figure}

The norms of the error vectors of the estimated distance-scaled translations and their local standard deviations are shown in Fig. \ref{blurError}. We use linear fitting to show their trends. The local standard deviations are calculated with the window size of 10\% of the total number of inliers.
For feature-based methods, if there are less than 2 inlier matchings in RANSAC, this pair is treated as an outlier. For the estimated pose, we apply the same local outlier rejection as Section \ref{section:NETWORKS} with the window size of 500. The final inlier rates of the network, SURF~\cite{bay2006surf}, ORB~\cite{rublee2011orb}, and FAST~\cite{trajkovic1998fast} are shown in Fig. \ref{blurError}. The network has the highest inlier rates partly because it does not rely on the number of matches so it can perform prediction on every image pair.
%There are two main observations from Fig. \ref{blurError}. First, 
Fig. \ref{blurError} shows that the network is most accurate with both datasets. Its performance is barely affected by the growing motion blur while feature-based methods more or less provide more noisy results. For the increasing disparity, the network is also least affected.

\subsection{Flight Dataset} \label{section:Flight Data}
% network is Table \ref{self-sup} 10
% \footnote{https://www.mynteye.com/pages/mynt-eye-d}
To obtain sensor data in flight, an MYNT EYE D1000-120 visual-inertial sensor is downward-facing mounted on an Eachine Wizard X220 FPV Racing Drone carrying an NVIDIA Jetson TX2. Its IMU measurements (200Hz) and monocular gray-scale images (30fps) with an exposure duration of 20ms are collected. The images are undistorted and transformed to have the same size and intrinsic matrix as the training set. The top left of Fig. \ref{fig1} shows an example. The camera's attitude is estimated by the Madgwick filter~\cite{madgwick2011estimation} using the IMU measurements. The ground-truth velocity is obtained from the OptiTrack motion tracking system at 120Hz. The first column of Table \ref{flight} shows the average and maximum distance-scaled translational velocity of 4 one-minute flights.
% with increasing speed.

\begin{table}[t]
	\centering
	
	\caption{Network and SURF Evaluated by Flight Dataset}
	\label{flight}
	\centering
	\begin{tabular} {p{23pt}<{\centering}p{195pt}<{\centering}}%{c{20pt}c{20pt}c{20pt}c{20pt}c{20pt}c{20pt}}
		\toprule  
		Sequence & RMSE (1e-2): network / SURF (orig.) / SURF (hist. equal.) \\
		\midrule 
		
		1(0.3,1.3)& (2.10, \underline{2.17}, \underline{2.01}) / (\underline{1.95}, \underline{2.17}, 2.40) / (\textbf{1.90}, \textbf{2.05}, \textbf{1.77})\\ 
		%			~& SURF& (2.18, 2.18, \textbf{2.65})\\  % (2.49, 2.50, 3.80) - 
		
		2(0.6,2.5)& (4.21, 4.57, \underline{3.65}) / (\underline{4.03}, \textbf{4.35}, 3.70) / (\textbf{3.91}, 4.36, \textbf{2.96})\\
		%			~& SURF& \\  % (2.49, 2.50, 3.80) - 
		
		3(1.2,3.2)& (\underline{5.44}, 5.87, \underline{5.03}) / (5.75, \underline{5.48}, 6.36) / (\textbf{5.20}, \textbf{5.25}, \textbf{4.34})\\ 
		%			~& SURF& \\  % (6.05, 6.44, 6.73) - 
		
		4(1.4,3.9)&(\underline{10.2}, \underline{9.48}, \underline{10.0}) / (15.1, 11.5, 28.3) / (\textbf{10.0}, \textbf{8.91}, \textbf{8.65})\\  
		%		4(1.4,3.9)&(\textbf{10.23}, \textbf{9.48}, \textbf{10.01}) / (15.10, 11.53, \textbf{28.33}) / (10.01, 8.91, \textbf{8.65})\\ % SURF failure_rate = 8.6435
		%			~& SURF& (11.75, 11.53, \textbf{11.76})\\ % (16.23, 12.87, 23.15) - 
		%		04& ORB& 14.4& (31.81, 26.19, 33.71)\\ 
		%		04& FAST& 45.5& (17.76, 14.67, 40.10)\\ 
		
		\bottomrule 
	\end{tabular}
	
\end{table}

\begin{figure}[!htbp]
	\centering
	\makebox{
		\includegraphics[scale=0.66]{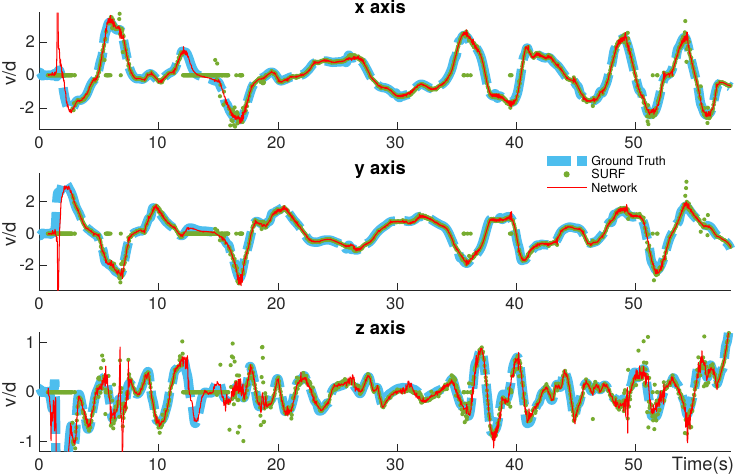}}
	\caption{The ratio of velocity to height (v/d) of the number 4 flight, expressed in the world frame. Both methods use original images.}
	\label{flight04}
\end{figure}

The results of the fastest flight are shown in Fig. \ref{flight04}. SURF's result is noisy in some parts of the flight mainly because of the big motion blur and scenes lacking texture. For 8.6\% of image pairs, SURF has less than 2 inlier matchings. We use zero vectors to show its results in this case. For the other 3 slower flights, SURF has enough matches all the time. The root mean square errors (RMSEs) of the distance-scaled velocity vector's components along the world frame's 3 axes are shown in Table \ref{flight}. When using original images, the network outperforms SURF more in faster flights where fewer points are detected. In histogram equalized images, more SURF points are detected and the accuracy is slightly higher than the network. The network performs better on original images than histogram equalized images since the images in the training set are without pre-processing.

\section{CONCLUSION}

In this article, we have shown that CNNs are suitable for ego-motion estimation of fast-moving MAVs equipped with a downward-facing camera. When flying fast, both motion blur and the visual disparity between subsequent images increase, which is handled better by a network than by traditional feature-based methods. Our investigation into the training of an ICSTN-based network shows that (1) it is better to take all blocks' prediction errors into account, (2) a larger receptive field that can be achieved by pyramidal images allows to estimate larger motions, (3) self-supervised learning based on the photometric error leads to better performance.

%and {\color{blue}(4) pose estimation network is a feasible onboard solution for real-time MAV navigation. For the next step, we will explore the uncertainty of the network's prediction to improve the visual-inertial ego-motion estimator.}

\section*{APPENDIX}

\subsection{Networks with Sharing Parameters among Blocks}

%The networks cannot be deeper under the constraint of inference speed and cannot be wider (more convolution kernels in each layer) under the constraint of model size. 
Sharing parameters among the blocks of an ICSTN-based network is a way to widen the network (have more convolution kernels in each layer) without enlarging the model size. We train the 6th network of Table \ref{icstn} and its variants by self-supervised learning. The results are shown in Table \ref{icstnShared}. Besides sharing all the parameters we also try only sharing the fully-connected layer (FC). 

The results show that sharing all the parameters and keeping the width of the network significantly reduce model size but hurt the accuracy a lot. The wider network having a slightly bigger model size fails to outperform the origin, either, shown in the 2nd row. Besides, it is slower (96Hz) than the original (101Hz). Sharing the fully-connected layer reduces the model size a little at the cost of the slight deterioration of accuracy.
An explanation to the results is that, when blocks have different parameters, the first block is trained to better handle bigger disparities and the last block focuses more on the smaller ones. Although sharing parameters widens the network, its enhancement to the model capacity of the block is less than the negative effects of weakening its specialization.

\begin{table}[!htbp]
	\centering
	\setlength{\tabcolsep}{0.52mm}{
		\caption{Networks with Shared Parameters}
		\label{icstnShared}
		\centering
		\begin{tabular} {p{23pt}<{\centering}p{68pt}<{\centering}p{135pt}<{\centering}}%{c{20pt}c{20pt}c{20pt}c{20pt}c{20pt}c{20pt}}
			\toprule  
			Share Params& Num. Params (original / wider)& EPE's Approx. Normal Distr.: std(1e-4) (original / wider)\\
			\midrule  
			None& 1.252M& (4.59, \textbf{4.04}, \textbf{5.94})\\ % 12-09-13:44 mavlab1
			All& 0.417M / 1.259M& (8.68, 8.42, 12.05) / (6.28, 5.96, 8.55)\\ % 12-09-13:54 godzilla0 / 12-09-13:56 godzilla1
			FC& 1.221M& (\textbf{4.44}, 4.11, 6.02)\\ % 12-09-13:49 mavlab0
			
			\bottomrule 
		\end{tabular}
	}
\end{table}

\begin{figure}[b]
	\centering
	\makebox{
		\includegraphics[scale=0.63]{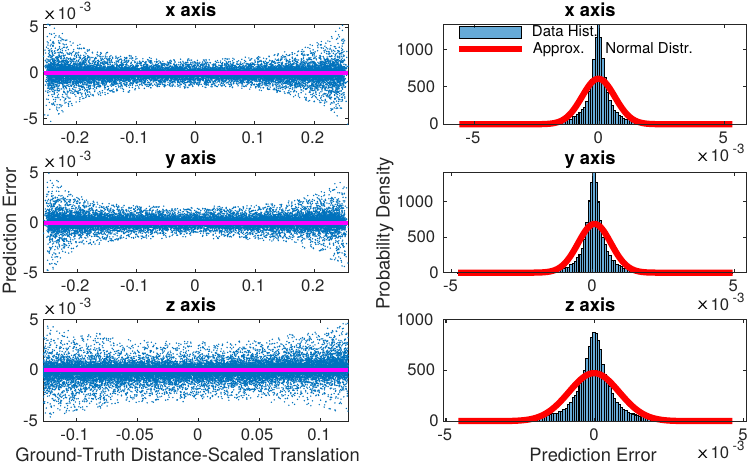}}
	\caption{Error distribution of the 6th network of Table \ref{icstn} trained in the self-supervised manner.}
	\label{distribution}
\end{figure}

\subsection{Error Distribution of Network's Prediction}\label{section:subsectionDistribution}

In order to better illustrate the performance of the networks, in Fig. \ref{distribution} we show the error distribution of the 6th network of Table \ref{icstn} trained in a self-supervised manner. Other networks' error distribution figures have similar shapes. After the outlier rejection described in Section \ref{section:METHODOLOGY}, its inlier rate is 85.2\%. From the left graph of Fig. \ref{distribution}, one can notice that the errors lie approximately unbiasedly close to zero and the predictions in the $z$-axis are noisier than the other 2 axes. For all the networks in this article the $z$-axis has worse predictions. This is also the case for most networks in~\cite{sanket2020prgflow}. It is possible that a standard CNN's ability to estimate scale variations is fundamentally limited (cf.~\cite{xu2014scale, van2017learning}). A deeper analysis of this issue is required.

%We think the difference from the standard normal distribution is mainly caused by the standard deviation growing along with the motion. 
Another phenomena worth noticing is that the network has noisier predictions with bigger translations. It is also shown in Fig. \ref{blurError}, the uncertainty of prediction grows with the amount of motion. And the prediction error does not perfectly normally distribute, as shown in the right graph of Fig. \ref{distribution}. Predicting the uncertainty of the pose prediction will be studied in future works.

%It is possible to train a deeper network with image pairs having even bigger motion to obtain error distribution closer to normal distribution. As for the current networks, 
%When we fuse the network predictions with other sensors for state estimation, it would be better to adaptively adjust the uncertainty according to the prior knowledge of motion (from the dynamic model or IMU integration) or using another network to predict the uncertainty than using the approximately normal distribution. 

\subsection{Public High-Speed Flight Dataset and Prior Pose}  \label{section:Public Dataset}

\begin{table}[b]
	\centering
	
	\caption{Evaluation by a Public Dataset of Fast MAV Flight}
	\label{UZH}
	\centering
	\begin{tabular} {p{42pt}<{\centering}p{169pt}<{\centering}}%{c{20pt}c{20pt}c{20pt}c{20pt}c{20pt}c{20pt}}
		\toprule  
		Sequence& RMSE (1e-1): Prior Pose Input / Zero Input\\
		\midrule 
		
		2 (6.97m/s)& (\textbf{10.76}, \textbf{11.81}, \textbf{7.53}) / (14.48, 12.63, 8.70)\\ 
		
		4 (6.55m/s)& (\textbf{9.94}, \textbf{14.64}, \textbf{6.80}) / (12.16, 14.77, 8.40)\\ 
		
		9 (11.23m/s)& (\textbf{9.70}, \textbf{12.78}, \textbf{10.55}) / (21.15, 15.07, 15.77)\\ 
		
		12 (4.33m/s)& (\textbf{7.74}, \textbf{8.44}, \textbf{6.30}) / (9.38, 8.85, 6.48)\\ 
		
		13 (7.92m/s)& (\textbf{9.86}, \textbf{19.08}, \textbf{6.56}) / (14.18, 19.13, 8.50)\\ 
		
		14 (9.54m/s)& (\textbf{33.06}, 22.27, \textbf{17.08}) / (36.75, \textbf{21.21}, 21.37) \\ 
		
		\bottomrule 
	\end{tabular}
\end{table}

\begin{figure}[b]
	\centering
	\makebox{
		\includegraphics[scale=0.168]{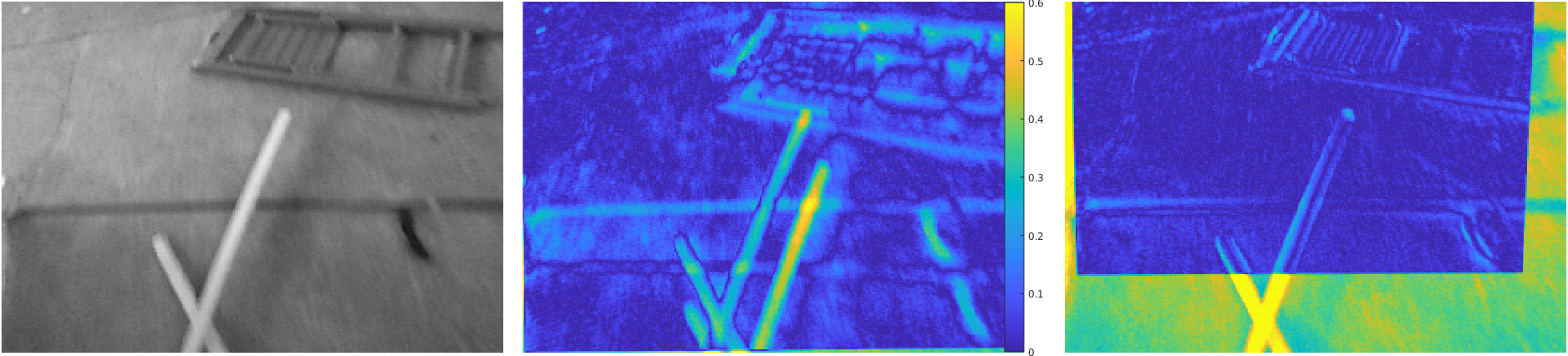}}
	\caption{The network's performance when the scene is not a perfectly planar surface. The left image is the first image. The middle one shows the photometric error after derotation and the right one shows the photometric error after the second image is warped by the network prediction.}
	\label{UZHseq2}
\end{figure}

\begin{figure}[t]
	\centering
	\makebox{
		\includegraphics[scale=0.63]{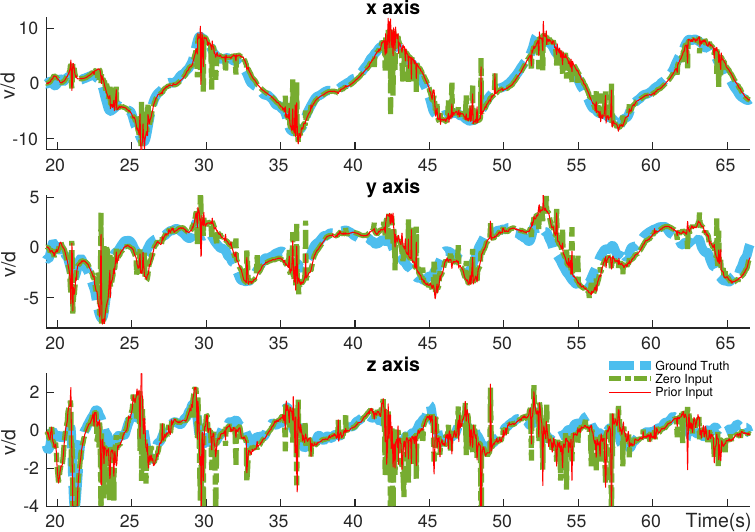}}
	\caption{The ratio of velocity to height expressed in the world frame. The network's results with and without the prior pose are compared using the number 2 indoor 45-degree downward-facing sequence of the UZH-FPV dataset.}
	\label{UZHSeq2}
\end{figure}

\begin{figure}[b]
	\centering
	\makebox{
		\includegraphics[scale=0.169]{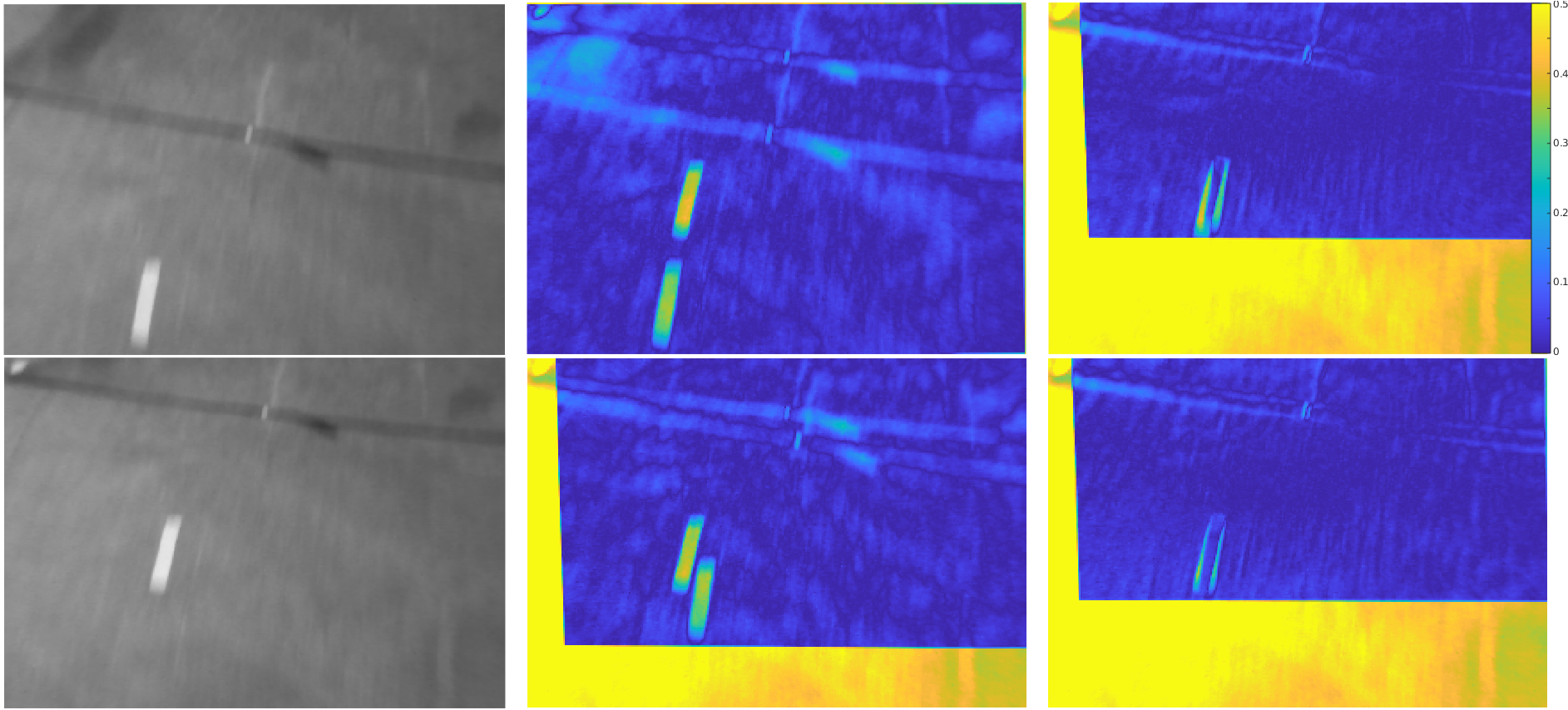}}
	\caption{Comparison between with and without the prior pose when the visual disparity is big. The left column shows the image pair; In the middle column, the upper image shows the photometric error after derotation and the lower one shows the photometric error of warping by the network prediction when a zero vector is the initial guess of translation; The upper image of the right column shows the photometric error of warping by the prior pose. And the lower one shows the photometric error of warping by network prediction when the prior pose is the input initial pose.}
	\label{UZHseq14}
\end{figure}

We evaluate the 4th network of Table \ref{self-sup} with the 6 indoor 45-degree downward-facing flight sequences that have public ground truth from the UZH-FPV~\cite{delmerico2019we} dataset. For this dataset, the distance to the ground is unknown. So we manually set the initial distance at the starting point of the ground truth data to make the RMSE smaller. The peak speed of each sequence and the networks' RMSEs are shown in Table \ref{UZH}. Fig. \ref{UZHseq2} shows that the network's performance when the scene is not a perfectly planar surface. Despite the network is disturbed by the low objects lying on the ground because of their rich visual textures, it still outputs relatively accurate predictions. There are other scenes that are not inside a single plane, such as high obstacles and strings. But since most of the time the ground plane takes up the majority of the image, the network has reasonable predictions. Another source of inaccuracy is that the attitude estimation from the Madgwick filter is less accurate because of the big acceleration during fast maneuvers.

In the previous parts of this article, the initial pose (shown in Fig. \ref{icstnFig}) of the network only contains the relative rotation. The translation is set to a zero vector. The network needs to deal with the whole visual disparity caused by translational motion. Because of the inertia of MAV, its translational velocity cannot change much during the sampling interval of the camera. So the translation vectors of temporally adjacent image pairs are similar. We add the predicted translation of the previous image pair as prior information to the initial pose and call it the prior pose. In this case, the disparity of the image pair warped by the initial pose gets smaller.

With the prior pose, the accuracy increases significantly and the predictions are less noisy, as shown in Table \ref{UZH} and Fig. \ref{UZHSeq2}. We think the reason is that the visual disparities of the image pairs in the UZH-FPV dataset are big enough for the network's accuracy to decrease. 
%The network's accuracy decreases with the increasing velocity as shown in Fig. \ref{blurError} and Fig. \ref{distribution}. 
Take the number 9 sequence as an example, the average absolute values of the difference between the network's input initial translation and output predicted translation in 3 axes significantly drop from 1.36e-1, 9.51e-2, and 6.46e-2 to 2.14e-2, 1.19e-2, and 1.50e-2 after using the prior pose. This means the network faces much smaller disparities. From Fig. \ref{UZHseq14} one can clearly see the big disparity when the MAV flies at around 9.2m/s close to the ground. Without the prior pose, the network decreases the disparity a lot but still not enough. By contrast, the disparities decrease a lot already after warping by the prior pose, which is easier for the network. The prior pose can be more accurate when other information sources of ego-motion (IMU, dynamics model, etc.) are available. The prior pose makes the network less demanded by big disparities, and thus makes it possible to reduce receptive fields and model sizes of the networks.

\subsection{CNN-based VIO for Real-Time Feedback Control}

For autonomous feedback control of an MAV using only onboard sensors and processors, we implement a CNN-based VIO for high-frequency ego-motion estimation. It is expanded from an EKF-based inertial attitude and velocity estimator~\cite{abeywardena2013improved} that also utilizes the linear drag model of quadrotor MAV. The network (4th of Table \ref{self-sup}) predictions and the LiDAR measurements are utilized in the measurement update of EKF.
	
As shown in Eq. \ref{eq1}, the translational vector is scaled by the height of the first image. This is also the case in the training of the networks. As for the VIO, we take the current image as the first image of the network input, and the previous image as the second. So the predicted distance-scaled translation is scaled by the current height. This avoids rerunning the EKF from the previous image's time on.
As the time interval of the image pair ($\Delta t$) is short, we assume the average velocity during $\Delta t$ approximately equals the instantaneous velocity of the current image. Then the measurement equation of the network prediction can be formulated as Eq. \ref{eq2}, where $h$ denotes the height of the camera, $\isotope[c]{\boldsymbol{v}}$ denotes the velocity vector expressed in the camera frame, and $\isotope[c]{\boldsymbol{t}}_{net,k}$ denotes the distance-scaled translation vector of the image pair predicted by the network. The prior pose ($\isotope[c]{\boldsymbol{t}}_{net,prior}$) is calculated from the estimated states. We input it to the network as the initial pose to reduce both amount and range of the motion that the network deals with. So we can ignore the varying error distribution of the network predictions over motion as shown in subsection \ref{section:subsectionDistribution}, and assume that the measurement noise of the network prediction ($\boldsymbol{n}_{net}$) is approximately Gaussian and the noise covariance matrix ($\boldsymbol{R}$) is constant.
		
\begin{equation}\label{eq2}
	\isotope[c]{\boldsymbol{t}}_{net,prior}= \frac{\Delta t \cdot \isotope[c]{\boldsymbol{v}}_{k|k-1}}{h_{k|k-1}},\ 	\isotope[c]{\boldsymbol{t}}_{net,k}=\isotope[c]{\boldsymbol{t}}_{net,prior} + \boldsymbol{n}_{net}
\end{equation}

The CNN-based VIO is implemented in C++ and communicates with the controller via Robot Operating System\footnote{http://wiki.ros.org/melodic} (ROS). In order to run the network implemented and trained in Python within the VIO, we use TorchScript and LibTorch from the PyTorch C++ API\footnote{https://pytorch.org/docs/stable/cpp\_index.html}. TorchScript generates the traced network model that can be loaded and run in C++ by LibTorch functions. 
%LibTorch converts variables from other types to Pytorch tensors and calls the network inference function. 
In flight, the average time cost of the network (4th of Table \ref{self-sup}) inference is around 12.8 milliseconds on the GPU of an NVIDIA Jetson TX2 \footnote{https://developer.nvidia.com/embedded/jetson-tx2}. Its MAXP\_CORE\_ARM power mode shows the highest inference speed of our network implementation. 
The camera's exposure duration is set to 10ms. Although the images look a little dark for bare eyes, the effect on the network's performance can be ignored. This highlights the network's generalizability. For control, a basic proportional-integral-derivative (PID)-based position and velocity controller runs on the TX2 as a ROS node. A Betaflight\footnote{https://betaflight.com/} flight controller is in charge of attitude control and connected with the TX2 via a universal asynchronous receiver-transmitter (UART). 

\begin{figure}[t]
	\centering
	\makebox{
		\includegraphics[scale=0.51]{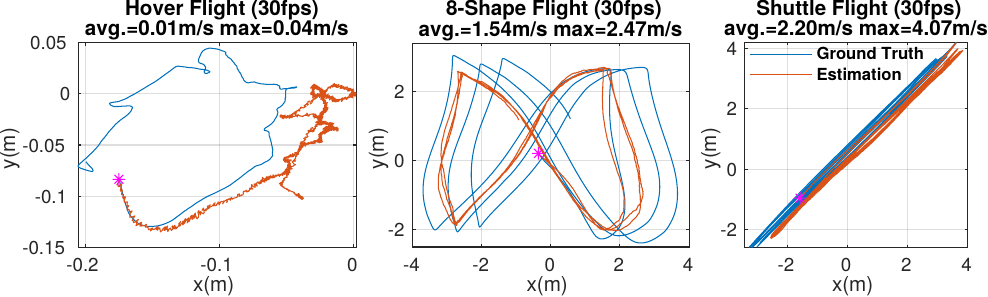}} % [scale=0.46]{trajectory3.pdf} [scale=0.65]{trajectory4.pdf}
	\caption{The ground-truth and estimated trajectories of hover flight (left), 8-shape flight (middle), and shuttle flight (right).}
	\label{trajectory}
\end{figure}

The MAV performs autonomous hover flight, eight-shape flight with changing heading and height, and high-speed shuttle flight between two waypoints, under using state estimation from the CNN-based VIO.
%We also tried 60-fps images for the shuttle flight, where around 1 in 8 images was dropped mainly because of the time-consuming of the network inference. 
The estimated and ground-truth trajectories for one-minute flights and their average and maximum speed are shown in Fig. \ref{trajectory}. The link to the flight video is shown in SUPPLEMENTARY MATERIALS. As far as we know, this is the first time that the pose estimation network's competence in autonomous feedback control of an MAV is demonstrated.
Note that the VIO shown here is basic. Unlike VIO solutions performing mapping, our VIO has no global correction. The network only predicts the relative pose of adjacent images. The trajectory is purely integrated from the network-corrected velocity estimation and thus suffers from drift over time. Taking the average velocity during camera sample interval as the instantaneous velocity and the constant noise covariance matrix of network prediction are two other sources of inaccuracy. There is space for further improvements.

%A TFmini\footnote{http://en.benewake.com/product/detail/5c345e26e5b3a844c472329c.html} LiDAR is mounted downward-facing and connected with the TX2 via Inter-Integrated Circuit (I$^2$C). Its measurement frequency is set to 50Hz.
%An MYNT EYE D1000-120\footnote{https://www.mynteye.com/pages/mynt-eye-d} visual-inertial sensor is mounted downward-facing on the MAV. In this work, it is set to outputs IMU measurements at 200Hz and monocular RGB images of size 640$\times$480 pixels at 30Hz. Images are then converted into gray-scale, undistorted, and transformed to have the same size and intrinsic matrix as the training images as described in subsection \ref{section:DatasetGeneration}. 

%When we increase the frame rate from 30 to 60, the time cost increase to 14.2 milliseconds.

\section*{SUPPLEMENTARY MATERIALS} \label{section:SUPPLEMENTARY}
The links to the videos demonstrating the network's performance and autonomous flights are \url{https://youtu.be/BMdh6dmLgrM} and \url{https://youtu.be/Uz9pNpn94jU}. The code developed for this work is open-source at \url{ https://github.com/tudelft/PoseNet_Planar}.
		
\section*{ACKNOWLEDGMENT}
The authors appreciate Ir. Nilay Y. Sheth for his supports in developing the MAV with GPU and collecting the datasets.

%%%%%%%%%%%%%%%%%%%%%%%%%%%%%%%%%%%%%%%%%%%%%%%%%%%%%%%%%%%%%%%%%%%%%%%%%%%%%%%%

\addtolength{\textheight}{-12cm}   % This command serves to balance the column lengths

\bibliographystyle{ieeetr}
\bibliography{reference.bib}

\end{document}